%% file: main.tex
\documentclass[11pt]{article}

\usepackage[margin=1in]{geometry}
\usepackage[utf8]{inputenc}
\usepackage[T1]{fontenc}
\usepackage{amsmath,amssymb}
\usepackage{graphicx}
\usepackage{booktabs}
\usepackage{array}
\usepackage{xcolor}

\definecolor{semdarkblue}{RGB}{18,52,98}   
\definecolor{semmidblue}{RGB}{47,94,148}   
\definecolor{segorange}{RGB}{214,108,26}   
\definecolor{semgray}{RGB}{110,110,110}    

\usepackage{caption}
\usepackage{subcaption}
\usepackage[colorlinks=true,linkcolor=semdarkblue,citecolor=semdarkblue,urlcolor=semmidblue]{hyperref}

\newcommand{\TBD}[1]{\textcolor{segorange}{\textit{pending}}}

\renewcommand{\thesection}{\Roman{section}}
\renewcommand{\thesubsection}{\Alph{subsection}}

\makeatletter
\renewcommand{\p@subsection}{\thesection-}
\renewcommand{\p@subsubsection}{\thesection-\thesubsection}
\makeatother

\title{\textbf{From Sound to Sense: An Anatomy of Semantic Supervision\\
for Speech-Driven Visual Grounding in Remote Sensing Imagery}}

\author{Zhongyu Pang$^{1}$\ \textcolor{segorange}{\textit{(co-authors
  pending)}}\\[4pt]
  \small\textit{$^{1}$College of Electronic Science, National University of
  Defense Technology, Changsha 410073, China}}
\date{October 4, 2026}

\makeatletter
\renewenvironment{abstract}{%
  \par\medskip\noindent\centerline{\bfseries Abstract}\par\bigskip\noindent\ignorespaces
}{%
  \par\medskip
}
\makeatother

\begin{document}

\maketitle

\begin{abstract}
Speech-driven visual grounding in remote sensing (RSVG) removes the typing
bottleneck of text-based interaction and is intended to preserve prosodic
intent, enabling
hands-free target acquisition in time-critical missions. The direct-audio
paradigm --- mapping raw speech straight to a bounding box and bypassing
automatic speech recognition (ASR) --- was recently adopted by RSTalker, but
what such grounders actually learn about \emph{language} remains unmeasured:
the only training signal is a box-regression loss, the reported
text-vs-audio comparison confounds backbone effects, and the standard test
split never exercises compositional generalization. We present
\textbf{SemEcho}, an instrumented anatomy of semantic supervision for
speech-driven RSVG. SemEcho grafts a frozen Whisper speech pathway onto a
VLTVG grounding host so that speech and text share one decoder, making the
paired text --- free by construction, since speech corpora in this task are
synthesized from DIOR-RSVG expressions --- a training-time teacher and a
fair, same-host text upper bound. On this instrument we dissect three
semantic injection channels (feature-mimicry distillation, word-level CTC
supervision, and two-stage ASR pretraining) with linear semantic probes, and
release \textbf{Audio-DIOR-R} (80,973 training clips; 10 timbres $\times$
$K{=}3$; Qwen3-TTS synthesis with online sox noise $\lambda\in[0,0.4]$) and
\textbf{Audio-DIOR-Comp}, a re-split holding out 17 attribute--category
combinations under a 2$\times$2 acoustic$\times$semantic protocol. Our
anatomy yields four findings. (i) Whether direct alignment beats a
domain-trained ASR cascade is decided by recognizer quality: against a weak
in-domain cascade (CTC 0.53) direct wins by 9.0 Acc@0.5 points with
transcription errors concentrated on spatial-relation words, but a strong
in-domain cascade (CTC 0.029) reaches the text upper bound (80.8 vs.\ 81.0)
and beats even the strongest direct model by 8.8 points --- the anti-ASR
premise holds only against
weak recognizers. (ii) All three word classes are already linearly
decodable from the frozen speech front-end with no grounding training
(category/attribute/relation 55.5/41.5/64.2\% vs.\ chances 5/12.5/20\%);
conventional grounding training collapses this inheritance (baseline
31.0/34.1/39.1), and stronger recipes progressively restore it --- the
relation-word dose response (39.1$\to$62.3, $+23.2$ points through v6;
74.8 at full strength, $+10.6$ above the frozen front-end) measures
recovery and enhancement of inherited semantics, not creation from
nothing. (iii)
Feature mimicry at conventional weights occupies 1--2\% of the training loss
and does not move Acc@0.5 ($-1.60$) --- yet is the most loss-efficient
restorer of inherited semantics ($+14.5$ relation points at that share) and,
via its orthogonal reservation, anchors acoustic identity through grounding
training (51.0\% pooled timbre decodability vs.\ ${\approx}$chance without;
75.0\% from the reserved residual);
word-level channels
restore relation-word decodability monotonically from the collapsed 39.1\%
to 62.3\% ($+23.2$ points). (iv)
The fair text upper bound is 72.97 Acc@0.5 / 62.31 mIoU. Scaling the speech
front-end to Whisper-Small and extending training under the full two-stage
recipe move the speech
route to 64.32 Acc@0.5 ($+11.04$ over the matched baseline; 62.81 on the
7,500-query test split) --- a joint effect of front-end capacity, training
schedule, and ASR pretraining that the matched-budget channels alone do not
produce --- closing 56\% of the 19.7-point text--speech gap; a
low-learning-rate training plateau persists while training loss keeps
falling, consistent with a structural limit (speech front-end word-identity
capacity) rather than under-training, though a longer-training control is
needed for confirmation. The gain persists under unseen acoustics
($+11.67$ vs.\ $+11.06$ on clean audio), and on the 17 held-out combinations
the recipe shows no compositional penalty over the supervision-free
baseline (41.59 vs.\ 40.11 Acc@0.5, statistically tied under a 20-epoch
schedule) --- compositional transfer remains open. All reported numbers use the full
3,829-query validation set with true-IoU evaluation. Code, the synthesis pipeline, the
re-split, and probes will be released.
\end{abstract}\noindent\ignorespaces


\input{intro}
\input{related}
\input{method}
\input{data}
\input{experiments}
\input{conclusion}

\bibliographystyle{unsrt}
\bibliography{refs}

\end{document}

%% file: intro.tex
\section{Introduction}
\label{sec:intro}

\subsection{Speaking to Earth-Observation Imagery}
\label{sec:speaking}

Remote sensing visual grounding (RSVG) localizes a target in an aerial image
from a natural-language query, serving as the interactive bridge between
mission intent and Earth-observation imagery~\cite{rstvg2023zhan}. In
time-critical deployments --- disaster response, traffic control, field
reconnaissance --- operators need hands-free, eyes-free interaction, and
spoken instructions carry prosodic cues (emphasis, urgency) that typed text
structurally discards.

\begin{figure}[t]
  \centering
  \includegraphics[width=\textwidth]{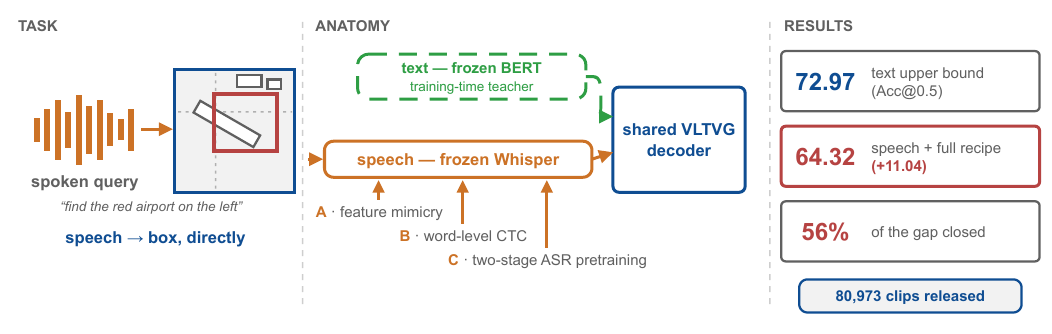}
  \caption{\textbf{SemEcho at a glance.} Spoken queries drive bounding-box
  prediction on remote-sensing imagery through a dual-stream host in which a
  frozen Whisper speech stream and a frozen BERT text stream share one
  decoder; three semantic distillation channels (A: feature mimicry,
  B: word-level CTC, C: two-stage ASR pretraining) inject language semantics
  at training time only. The full recipe reaches 64.32 Acc@0.5 ($+11.04$ over
  the matched baseline), closing 56\% of the 19.7-point text--speech gap;
  80,973 clips are released.}
  \label{fig:teaser}
\end{figure}

Audio-driven RSVG~\cite{rstalker2026liu} answers this need: raw speech is
mapped directly to a bounding box, bypassing the ASR cascade whose
transcription errors propagate catastrophically in the remote-sensing
vocabulary (e.g., ``track''$\,\to\,$``truck''). We fully endorse the
direct-audio premise --- indeed, this paper provides its first
\emph{controlled} quantitative test within our knowledge
(Section~\ref{sec:directcascade}); with a single prior paper in this
setting, primacy claims are necessarily provisional. Our
concern is a different one: \textbf{what does a direct-audio grounder know
about language, and where does that knowledge come from?}

\subsection{Four Blind Spots in the Current Paradigm}
\label{sec:blindspots}

We revisit RSTalker~\cite{rstalker2026liu}, the state of the art in
audio-driven RSVG, and identify four gaps that follow from what the original
publication reports. All four are checkable against it.

\begin{enumerate}
  \item \textbf{No reported semantic supervision.} The published objective
  consists of $\ell_1$ and GIoU box-regression losses (plus deep supervision
  of the same form); no semantic term is reported. Nothing in the objective
  requires the model to know
  \emph{what the sentence means}; the query generator is a purely acoustic
  gating-and-pooling module. The speech-to-box mapping is free to fit
  statistical acoustic--spatial correlations of the 10-timbre noisy training
  corpus.
  \item \textbf{A confounded modality comparison.} The reported
  text-vs-audio gap compares a ResNet-101+BERT grounding model
  (MGVLF~\cite{rstvg2023zhan}) against a Swin-T+Whisper
  model~\cite{rstalker2026liu}. No same-backbone, same-decoder text control
  exists, so the cost of the speech modality cannot be separated from the
  benefit of a stronger backbone.
  \item \textbf{Unmeasured prosody claims.} Prosodic disambiguation
  (``stressing \emph{red} vs.\ \emph{ship}'') motivates the entire paradigm,
  yet no experiment manipulates or measures prosodic sensitivity.
  \item \textbf{A compositional blind spot in the test protocol.} 277 of
  the 278 (attribute, category) combinations appearing in the Audio-DIOR
  test split also appear in training under a 30-word attribute lexicon ---
  the single exception is (``brown'', ``ground track field'') --- so the
  strictly compositional (zero-shot combination) subset is effectively
  empty (Section~\ref{sec:blindspot}). Acoustic robustness is engineered by
  data augmentation, but \emph{semantic} generalization is never exercised.
\end{enumerate}

Gaps 1--3 are properties of the training signal; gap 4 is a property of the
measurement. Both must be fixed before any claim of ``the audio grounder
understands language'' is testable.

\subsection{This Paper: An Anatomy, Not a New Module}
\label{sec:anatomy}

We therefore approach the problem as an \textbf{anatomy study}. We build an
instrumented dual-stream grounder on a published, reproducible host
(VLTVG~\cite{vltvg2022}, whose DIOR-RSVG text anchor is 75.79 Acc@0.5), in
which a frozen BERT text stream and a frozen Whisper speech stream are
projected into the \emph{same} token space and decoded by the \emph{same}
multi-stage decoder. Because every speech clip in this task is synthesized
from a DIOR-RSVG expression, the paired text is free by construction; the
text stream doubles as (i) a task-grounded training-time teacher and (ii) the
missing fair text upper bound of~\cite{rstalker2026liu}.

Onto this instrument we mount a measurement suite and three semantic
injection channels, then dissect:

\begin{itemize}
  \item \textbf{Linear semantic probes} (Section~\ref{sec:probes}) that
  measure what the speech representation linearly encodes: which
  category/attribute/relation word was spoken, the spoken bag-of-words, and
  sentence-level discriminability via speech$\,\to\,$text retrieval.
  \item \textbf{Direct vs.\ cascade} (Section~\ref{sec:directcascade}): the
  first controlled comparison between direct speech-to-box alignment and a
  \emph{strong in-domain} ASR cascade (a CTC recognizer trained on the same
  81k remote-sensing expressions), on identical samples.
  \item \textbf{Three injection channels}
  (Sections~\ref{sec:channels},~\ref{sec:mainresults}, and~\ref{sec:gradient}):
  feature-mimicry distillation from the text teacher; word-level CTC
  supervision; and two-stage ASR pretraining --- compared under identical
  hosts, data, and budgets, with a loss-share accounting that explains
  \emph{why} one of them fails.
\end{itemize}

\subsection{Contributions}
\label{sec:contributions}

\begin{enumerate}
  \item \textbf{Audio-DIOR-R}, an open, recipe-faithful re-synthesis of the
  unpublished Audio-DIOR corpus: DIOR-RSVG expressions $\to$
  Qwen3-TTS-12Hz-1.7B-VoiceDesign with 10 timbre profiles $\times\,K{=}3$
  variants and an online sox noise library
  ($\lambda \sim \mathcal{U}[0,0.4]$); 80,973 training clips
  (Section~\ref{sec:audiodiorr}).
  \item \textbf{Evaluation methodology}: a compositional blind-spot analysis
  of the standard split; \textbf{Audio-DIOR-Comp}, a C-GQA-style re-split
  holding out 17 attribute--category combinations (1,284 comp-test queries)
  under a 2$\times$2 acoustic$\times$semantic matrix; the \textbf{linear
  semantic probe} protocol that turns ``does the audio grounder understand
  language?'' into measurable quantities
  (Sections~\ref{sec:probes} and~\ref{sec:data}); and the controlled
  measurement showing the recipe's gains localize to seen compositions
  (Section~\ref{sec:compacoustic}).
  \item \textbf{A controlled anatomy of semantic supervision}: the same-host
  text upper bound (72.97 Acc@0.5 / 62.31 mIoU on the full validation set);
  direct-vs-cascade quantification (direct $+9.0$ over a weak cascade; a
  strong cascade $+8.8$ over the strongest direct model --- recognizer
  quality decides); and
  a three-channel dissection with an honest account of a negative result ---
  feature mimicry is a behavioral no-op at conventional weights (1--2\% loss
  share, $-1.60$ Acc@0.5), yet the most loss-efficient representation-level
  restorer ($+14.5$ relation points) and the only channel that anchors
  acoustic identity through grounding training, while word-level channels
  restore relation-word
  decodability monotonically from the collapsed 39.1
  (39.1$\,\to\,$53.6$\,\to\,$62.3; the frozen front-end's inheritance line
  is 64.2, which the full-strength v7 crosses at 74.8)
  --- and a full-strength recipe (Whisper-Small front-end $+$ two-stage ASR
  pretraining $+$ layer-weighted summation $+$ decoder-initialized connector)
  that moves the speech route to 64.32 Acc@0.5 ($+11.04$ over the matched
  baseline; 62.81 on the 7,500-query test split), closing 56\% of the
  text--speech gap (Sections~\ref{sec:mainresults}--\ref{sec:gradient}).
  \item \textbf{Mechanism findings} with guidance for future audio grounders:
  word-class decodability is inherited from the frozen speech front-end and
  \emph{collapsed} by conventional grounding training (frozen
  55.5/41.5/64.2 vs.\ baseline 31.0/34.1/39.1) --- recipes are judged by how
  much inheritance they preserve or restore, and only sentence
  discriminability is learned from nothing; ASR transcription
  errors concentrate on exactly the spatial-relation words that grounding
  needs most (a recognizer that recovers them erases the cascade's deficit
  entirely); and injection channels must be ranked by the share of the
  training loss they actually command, not by loss design.
\end{enumerate}

%% file: related.tex
\section{Related Work}
\label{sec:related}

\subsection{Visual Grounding in Remote Sensing}
\label{sec:rsvg-related}

Specialized RSVG models evolved from two-stage pipelines to one-stage
transformers: TransVG~\cite{transvg2021deng}, VLTVG~\cite{vltvg2022}, and
MGVLF~\cite{rstvg2023zhan}; MLLM-based grounders scale to 7B+ parameters.
All of these assume \emph{textual} queries. VLTVG~\cite{vltvg2022} is, to our
knowledge, the strongest published openly reproducible architecture on
DIOR-RSVG (75.79 Acc@0.5 on the test split), which is why we adopt it as the
host of our anatomy (Section~\ref{sec:host}).

\subsection{Speech-Driven Grounding and Audio-Visual Alignment}
\label{sec:speech-grounding}

Audio-visual grounding spans natural images, video, and medical
imaging~\cite{avgsurvey2024zhu}. RSTalker~\cite{rstalker2026liu} opened the
speech-conditioned grounding
setting with a direct speech-to-box pathway (Swin-T~\cite{swin2021liu} +
Whisper + a speech-conditioned query generator). STBridge~\cite{stbridge2023li}
addresses noise-tolerant speech-referred video object segmentation in natural
scenes, where the referent itself may emit sound; remote sensing differs in
that targets are silent and instructions are human spoken commands. No prior
work measures the \emph{semantic content} of the audio pathway of a grounder,
which is the focus here.

\subsection{Injecting Language Semantics Without an ASR Cascade}
\label{sec:semantics-injection}

End-to-end spoken language understanding maps acoustics to semantics
directly, and speech-LLM pipelines such as SLAM-LLM~\cite{slamllm2024} couple
a speech encoder to an LLM for generative tasks. Closest to our injection
channels, DiVA~\cite{diva2024held} distills an LLM's text pathway into a
voice assistant without in-domain speech data, using exactly the paired text
that TTS synthesis provides for free --- the same trick we exploit for
grounding, but our target is a \emph{localization} head, where semantics must
survive down to box coordinates, and our diagnosis (Section~\ref{sec:gradient})
shows that naive feature mimicry is a numerical no-op in this setting.
Speech$\,\to\,$text embedding alignment has also been explored for medical
imaging (SpeechCT-CLIP~\cite{speechctclip}: voice-controlled CT volume
localization); all of these target global
utterance embeddings or generation, none a referring localizer.

\subsection{Knowledge Distillation and Orthogonal Subspaces}
\label{sec:kd-orthogonal}

Classic KD~\cite{kd2015hinton} and its attention/relational variants transfer
soft targets; orthogonal decomposition appears in
OV-OrthKD~\cite{ovorthkd2024} for open-vocabulary audio-visual event
localization. We repurpose
orthogonality for \emph{modality disentanglement} (semantics vs.\
timbre/prosody) inside a single token representation, with a speaker probe
measuring the disentanglement (Section~\ref{sec:channelA}). Our
loss-share accounting (Section~\ref{sec:gradient}) adds a practical
criterion that KD-based multimodal work rarely reports: the fraction of the
training loss an auxiliary loss actually commands. Our controls add a second
distinction the KD literature rarely measures: because the student may
already \emph{inherit} the target semantics from its frozen front-end,
distillation can act as \textbf{preservation} --- anchoring inherited
structure against task pressure --- rather than creation;
Section~\ref{sec:profile} quantifies both roles.

\subsection{Compositional Generalization Protocols}
\label{sec:compositional}

C-GQA~\cite{cgqa2022} introduced held-out attribute--category compositions
for grounding on natural images. Audio-DIOR-Comp (Section~\ref{sec:comp})
ports this idea to speech-driven RSVG and combines it with the acoustic axis
(seen vs.\ held-out timbres and noises) into a 2$\times$2 protocol --- to our
knowledge the first compositional protocol for speech-driven RSVG, in what
remains a one-prior-paper setting.

\nocite{remoteclip2024}

%% file: method.tex
\section{Methodology}
\label{sec:method}

\subsection{Problem Formulation}
\label{sec:formulation}

Given an image $\mathbf{I} \in \mathbb{R}^{3\times H\times W}$ and raw speech
$\mathbf{A} = \{\mathbf{a}_1, \ldots, \mathbf{a}_L\}$, predict the target box
\begin{equation}
  \mathbf{B}^{*} = \arg\max_{\mathbf{B}}\;
  P\!\left(\mathbf{B} \mid \mathbf{I}, \mathbf{A}; \theta\right).
  \label{eq:objective}
\end{equation}
Paired text $T$ exists for every speech clip by construction (Audio-DIOR is
synthesized from DIOR-RSVG triplets $(\mathbf{I}, T, \mathbf{B})$~\cite{rstvg2023zhan,rstalker2026liu});
our teacher and all injection channels use $T$ at \textbf{training time only}.

\subsection{The Dual-Stream Host}
\label{sec:host}

\begin{figure}[t]
  \centering
  \IfFileExists{figs/fig1_framework.pdf}{%
    \includegraphics[width=0.98\textwidth]{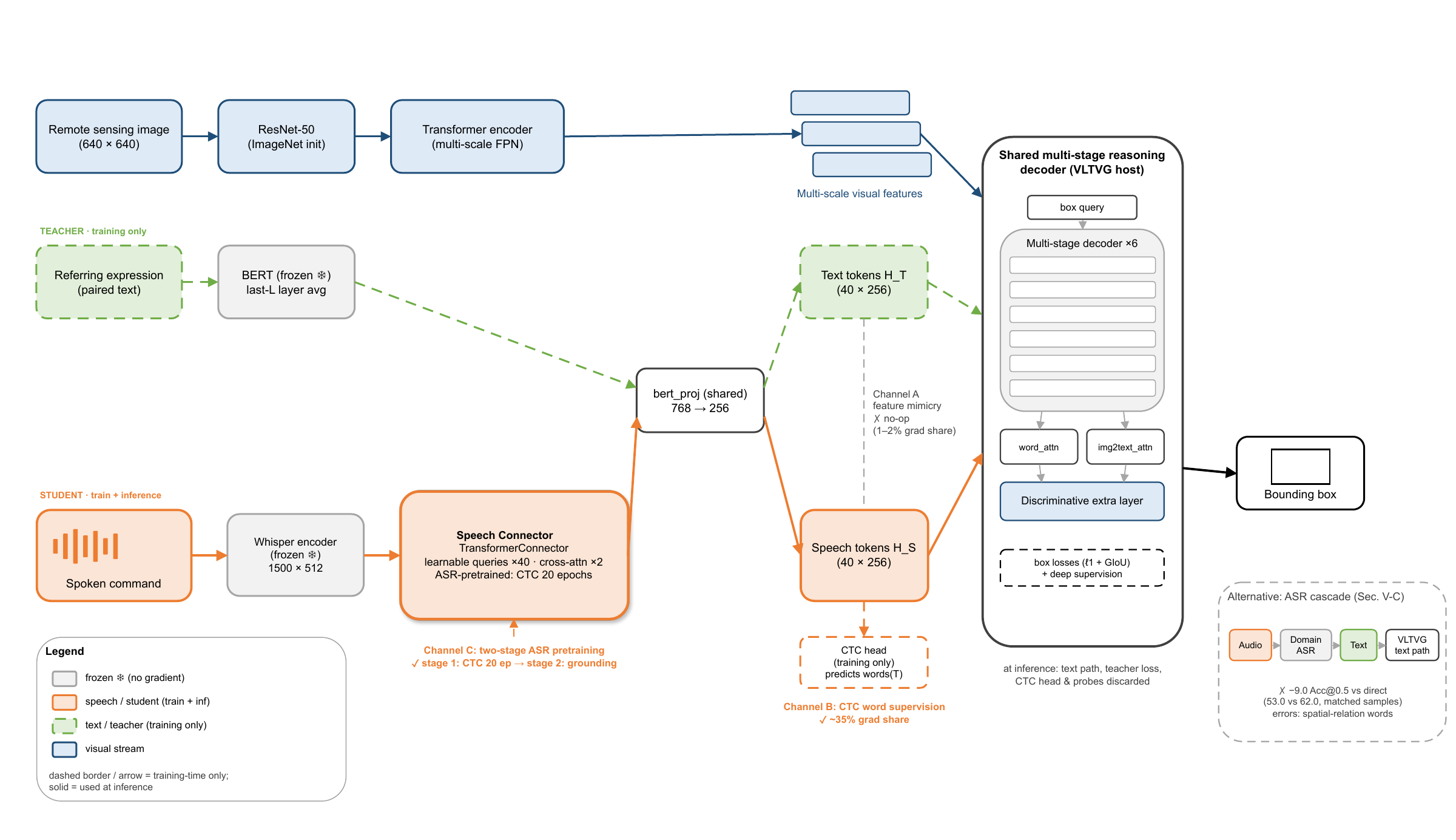}}{%
    \fbox{\parbox[c][5.2cm][c]{0.92\textwidth}{\centering
      \textcolor{semgray}{\itshape Placeholder for
      Fig.~\ref{fig:framework} (\texttt{fig1\_framework.pdf}): the VLTVG
      dual-stream host with the frozen BERT text stream, the frozen Whisper
      speech pathway and speech adapter, the shared \texttt{bert\_proj}
      projection, the shared multi-stage decoder, and the three semantic
      injection channels. The figure is generated by a separate build step
      and will be placed at \texttt{figs/fig1\_framework.pdf}.}}}}
  \caption{The SemEcho dual-stream host. A frozen Whisper-Base.en encoder
  feeds a lightweight speech adapter whose 40 projected tokens enter the
  \emph{same} \texttt{bert\_proj} projection and the \emph{same} VLTVG
  multi-stage decoder as the frozen BERT text stream; the paired text is free
  by construction because every clip is synthesized from a DIOR-RSVG
  expression. Three semantic injection channels (feature mimicry, word-level
  CTC, two-stage ASR pretraining) attach to the speech stream and operate at
  training time only.}
  \label{fig:framework}
\end{figure}

We graft a speech stream onto VLTVG~\cite{vltvg2022}, a progressive
query-modulated grounding decoder whose language interactions are
concentrated in its token stream: the decoder maintains a single box query
and performs word-to-image and image-to-word cross-attention
(\texttt{word\_attn} / \texttt{img2text\_attn}) against a 40-token language
feature map. Any stream that produces a $(40, d)$ token map can therefore
drive the identical decoder --- the property our anatomy exploits
(Fig.~\ref{fig:framework}).

\paragraph{Text stream (teacher; training only).} Frozen
BERT~\cite{bert2019devlin} encodes $T$; the last-$L$ layer token features are
averaged ($L$ as in the official VLTVG configuration) and projected by a
shared linear \texttt{bert\_proj} (768$\,\to\,$256) to word tokens
$\mathbf{H}_T \in \mathbb{R}^{40\times 256}$.

\paragraph{Speech stream (student; training + inference).} The frozen
Whisper-Base.en encoder~\cite{whisper2022radford} maps the log-mel waveform
to 1,500 frames (512-d). A lightweight \textbf{speech adapter} compresses
them to 40 tokens:
\begin{equation}
  \mathbf{z}_s = \mathrm{LN}\!\left(\mathrm{Pool}_{40}\!\left(
  \mathrm{GELU}\!\left(\mathrm{Conv}_{k3}\!\left(\mathrm{GELU}\!\left(
  \mathrm{Conv}_{k5,s4}(\mathbf{S})\right)\right)\right)\right)\right)
  \in \mathbb{R}^{40\times 768},
  \label{eq:adapter}
\end{equation}
followed by the \textbf{same} \texttt{bert\_proj} to
$\mathbf{H}_S \in \mathbb{R}^{40\times 256}$; the box query then attends to
$\mathbf{H}_S$ in the shared decoder exactly as it attends to $\mathbf{H}_T$.

\paragraph{Feature-scale normalization is not optional.} With a randomly
initialized adapter, the projected speech tokens had std $0.068$ ---
one-sixth of the BERT stream's $0.391$ --- which collapses the decoder's
word-attention to near-uniform and yields pathological training; adding the
output LayerNorm in Eq.~\eqref{eq:adapter} restores std $0.575$, on par with
the text stream. We also note the speech stream must use an all-valid token
mask: the 40 pooled tokens are densely global and cannot inherit BERT's
padding mask.

\paragraph{Inference profile.} At test time the speech path runs alone; the
text stream, teacher supervision, CTC head, and probes are discarded.
Inference cost is bit-identical to a speech-only grounder of the same host ---
all semantic machinery is training-time (Section~\ref{sec:cost}).

This host gives the anatomy its reference points:
\begin{itemize}
  \item \textbf{Fair text upper bound} --- the text stream fine-tuned on
  DIOR-RSVG fills the missing same-backbone control
  of~\cite{rstalker2026liu} (Table~IV there compares MGVLF~\cite{rstvg2023zhan}
  against its own speech model);
  \item \textbf{Task-grounded teacher} --- in the mimicry channel the teacher
  is not a frozen embedding but the full text pathway \emph{with box
  supervision on the same image}, so a teacher token representation is one
  that provably localizes the target.
\end{itemize}

\subsection{Three Semantic Injection Channels}
\label{sec:channels}

The host receives no language supervision by itself: the speech path is
trained purely by box losses. We dissect three channels of increasing
directness through which language semantics can be injected into the speech
representation. They are mutually comparable --- same host, same data, same
budget --- and are evaluated head-to-head in Section~\ref{sec:mainresults}.

\subsubsection{Channel A: Feature-Mimicry Distillation}
\label{sec:channelA}

The original SemEcho design distills the speech tokens $\mathbf{H}_S$ toward
the (stop-gradient) teacher tokens $\mathbf{H}_T$ at three levels, plus an
orthogonality constraint that reserves a residual subspace for acoustic
identity.

\paragraph{Query/token level ($L_q$).} Let
$\mathbf{P} \in \mathbb{R}^{256\times 64}$ be a learned projection with
L2-normalized columns, defining a 64-d \emph{semantic subspace}. Distill
token-wise inside it:
\begin{equation}
  L_q = \frac{1}{40}\sum_{t}\Bigl[1-\cos\!\left(
  \mathbf{P}^{\top}\mathbf{h}^{\mathrm{S}}_t,\,
  \mathbf{P}^{\top}\mathbf{h}^{\mathrm{T}}_t\right)\Bigr]
  + \left\lVert \mathbf{P}^{\top}\mathbf{H}_S
  - \mathbf{P}^{\top}\mathbf{H}_T \right\rVert_2^2.
  \label{eq:lq}
\end{equation}

\paragraph{Attention level ($L_{\mathrm{attn}}$).} For every decoder
cross-attention map $\mathbf{a} \in \mathbb{R}^{1\times 40}$ (head-averaged;
\texttt{word\_attn} and \texttt{img2text\_attn}), match the student to the
teacher at temperature $\tau=2$:
\begin{equation}
  L_{\mathrm{attn}} = \frac{1}{|\mathcal{M}|}\sum_{m\in\mathcal{M}}
  \mathrm{KL}\!\left(\mathrm{softmax}\!\left(\mathbf{a}^{\mathrm{T}}_m/\tau\right)
  \,\Vert\,
  \mathrm{softmax}\!\left(\mathbf{a}^{\mathrm{S}}_m/\tau\right)\right).
  \label{eq:lattn}
\end{equation}

\paragraph{Global level ($L_{\mathrm{sem}}$).} The mean-pooled speech
embedding is aligned to the teacher's pooled text embedding with
bidirectional InfoNCE~\cite{cpc2018oord} at temperature $0.07$.

\paragraph{Orthogonal prosody preservation ($L_{\mathrm{orth}}$, speaker
probe).} Distillation acts only on $\mathbf{P}^{\top}\mathbf{h}$; the
residual $\left(\mathbf{I}-\mathbf{P}\mathbf{P}^{\top}\right)\mathbf{h}$ is
free. Column orthonormality is enforced by
$\lVert \mathbf{P}^{\top}\mathbf{P} - \mathbf{I} \rVert_F^2$ regularization,
and a linear probe classifies the synthesis timbre (10 classes) from the
pooled residual, measuring whether acoustic identity is preserved in the
reserved subspace.

The total objective adds the teacher's own box loss (task grounding):
\begin{align}
  L^{(\mathrm{A})} = {}& L_{\mathrm{box}} + L_{\mathrm{box}}^{\mathrm{T}}
  \nonumber\\
  &+ w(t)\left(\lambda_q L_q + \lambda_{\mathrm{attn}} L_{\mathrm{attn}}
  + \lambda_{\mathrm{sem}} L_{\mathrm{sem}} + \lambda_{\mathrm{orth}}
  L_{\mathrm{orth}}\right)
  + \lambda_{\mathrm{sp}}\,\mathrm{CE}_{\mathrm{probe}},
  \label{eq:channelA}
\end{align}
with ramp $w(t)$ over 3 warm-up epochs and
$\lambda=(1,\,0.5,\,0.5,\,0.01,\,0.1)$. Section~\ref{sec:gradient} shows that
despite this careful design, channel A commands only 1--2\% of the training
loss at these weights and is a numerical no-op for grounding.

\subsubsection{Channel B: Word-Level CTC Supervision}
\label{sec:channelB}

Channel A constrains \emph{geometry} (where tokens sit in representation
space) but never says \emph{what words were said}. Channel B adds an
explicit word-level objective: a linear ASR head over the projected speech
tokens predicts the BERT word sequence of the source expression with a CTC
loss~\cite{graves2006ctc} --- alignment-free over the 40 tokens, blank at
index 0:
\begin{equation}
  L_{\mathrm{ctc}} = \mathrm{CTC}\!\left(
  \mathrm{softmax}\!\left(\mathbf{W}_{\mathrm{asr}}\mathbf{H}_S\right),\;
  \mathrm{words}(T)\right).
  \label{eq:lctc}
\end{equation}
$\mathbf{W}_{\mathrm{asr}}\in\mathbb{R}^{256\times|V|}$
($|V| = 30{,}522$ BERT words) is training-only. At weight $\lambda=1.0$ this
term commands ${\sim}35\%$ of the training loss in the initial iterations
(3--8\% once the CTC loss converges; Section~\ref{sec:gradient}) --- still
several times the share of channel A at any point in training.

\subsubsection{Channel C: Two-Stage ASR Pretraining}
\label{sec:channelC}

Channel B asks the adapter to learn word recognition \emph{concurrently} with
grounding. Channel C front-loads it, DiVA-style~\cite{diva2024held}:
\begin{itemize}
  \item \textbf{Stage 1 (ASR pretraining).} CTC-only training of the speech
  front-end (Eq.~\eqref{eq:adapter} for the conv adapter, or the transformer
  connector of Section~\ref{sec:mainresults}'s full-strength rows) $+$ ASR
  head on all 80,973 training clips --- no image forward, ${\sim}3\times$
  faster per step --- with AdamW lr $2\times10^{-4}$ and batch 16 (batch 8
  for v3). Three generations were trained as the recipe evolved, and we
  report each run with the generation it actually loaded
  (Table~\ref{tab:asrgen}).
  \item \textbf{Stage 2 (grounding).} Initialize the grounding run's speech
  front-end from Stage 1 and train the full host with a reduced CTC anchor
  $\lambda=0.3$.
\end{itemize}

\begin{table}[htbp]
  \centering\small\setlength{\tabcolsep}{3pt}
  \caption{Two-stage ASR pretraining: three generations. Each grounding run
  is reported with the generation it actually loaded.}
  \label{tab:asrgen}
  \begin{tabular}{@{}llrrrp{3.4cm}@{}}
    \toprule
    Generation & Front-end & Epochs & Final CTC & Wall-clock & Loaded by \\
    \midrule
    v1 (\texttt{adapter.pt}) & conv adapter & 5 & 0.53 & ${\sim}1.4$\,h &
    v6; the cascade recognizer of Section~\ref{sec:directcascade} \\
    v2 (\texttt{connector\_v2.pt}) & transformer connector & 20 & 0.006 &
    ${\sim}5.4$\,h & v7 \\
    v3 (\texttt{connector\_v3.pt}) & transformer $+$ decoder-init & 15 &
    0.029 & ${\sim}11$\,h & v8 / v8\_ft2 \\
    \bottomrule
  \end{tabular}
\end{table}

In every generation the loss falls by 2--4 orders of magnitude (to 0.53 /
0.006 / 0.029): the front-end first masters transcribing the remote-sensing
word sequence.

The three channels form the ablation axis of this paper: \emph{none}
(speech-only baseline) $\to$ \emph{mimicry} (A) $\to$ \emph{word-level, from
scratch} (B) $\to$ \emph{word-level, pretrained} (C).

\subsection{Linear Semantic Probes}
\label{sec:probes}

Accuracy alone cannot distinguish ``understands language'' from ``fits
acoustic--spatial correlations''. We therefore probe the trained speech
representation $\mathbf{H}_S$ (mean-pooled 40 tokens after
\texttt{bert\_proj}) with lightweight linear heads --- the protocol
popularized for speech representations by SUPERB~\cite{superb2021yang} ---
on a fixed 500-query
validation subset, identical across all compared models, with timbre/noise
balanced so probes cannot cheat through synthesis artifacts:
\begin{enumerate}
  \item \textbf{Word-class probes.} Multi-class probes for (i) the 20 target
  categories, (ii) the attribute word (9-color lexicon) on the
  attribute-bearing subset, and (iii) the spatial relation word (8-word
  lexicon) on the relation-bearing subset. Chance is 5.0\% for the 20-way
  category probe; chances for (ii)/(iii) are reported with each result and
  are the \emph{empirical class frequencies of the probe subset}, not
  $1/K$. Every word-class probe claim is reported against two nulls ---
  chance and untrained controls of the frozen front-end (frozen features,
  random adapter, random projection; Section~\ref{sec:profile}).
  \item \textbf{Bag-of-words probe.} A multi-label probe over the top-40
  frequent words (frequency $\ge 100$) measures raw word-identity information
  --- the ``which words were said'' ceiling that acoustics alone should
  nearly solve.
  \item \textbf{Speech$\,\to\,$text retrieval.} Cosine retrieval of the
  paired expression among a pool, reporting R@1 / R@10
  (chance ${\approx}0.2\%$) --- sentence-level discriminability.
  \item \textbf{Speaker probe.} Timbre classification from the orthogonal
  residual (Section~\ref{sec:channelA}) --- prosody/timbre preservation.
  \item \textbf{Bucketed localization.} Acc@0.5 on query buckets split by
  linguistic content: relation-bearing, attribute-bearing, and simple
  (category-only) queries --- connecting decodability to grounding behavior.
\end{enumerate}

Together these probes decompose the text--speech gap into \emph{word
identity} (acoustic), \emph{word class} (semantic), \emph{sentence
discriminability} (compositional), and \emph{localization} (task) components.

%% file: data.tex
\section{Audio-DIOR-R and the Evaluation Protocols}
\label{sec:data}

\subsection{Audio-DIOR-R: An Open Re-Synthesis}
\label{sec:audiodiorr}

The official Audio-DIOR corpus is unpublished; its synthesis recipe, however,
is fully specified in~\cite{rstalker2026liu}. We replicate the recipe
faithfully (the corpus itself remains unpublished):
DIOR-RSVG expressions $\to$ text normalization following~\cite{rstalker2026liu}
(preserving category, attribute, count, and spatial-relation words) $\to$
\textbf{Qwen3-TTS-12Hz-1.7B-VoiceDesign}~\cite{qwen3tts2025} with
\textbf{10 timbre profiles $\times\,K{=}3$ variants} $\to$ online mixing of a
\textbf{sox-synthesized noise library} (wind/engine/mechanical) with mixing
ratio $\lambda \sim \mathcal{U}[0,0.4]$ $\to$ the original train/val/test
assignment. Test-Acoustic uses 2 held-out timbres plus unseen noise clips.
Table~\ref{tab:splits} summarizes the corpus.

\begin{table}[t]
  \centering\small\setlength{\tabcolsep}{5pt}
  \caption{Audio-DIOR-R splits. Train, val, Test-Clean, and Test-Acoustic
  audio are synthesized (Test-Acoustic: 22,493/22,500 clips; the 7 missing
  clips affect evaluated totals by at most 0.1 point).}
  \label{tab:splits}
  \begin{tabular}{@{}lrrlp{4.6cm}@{}}
    \toprule
    Split & Queries & Clips & Hours & Notes \\
    \midrule
    Train & 26{,}991 & 80{,}973 & 53.5\,h & 10 timbres $\times\,K{=}3$ +
    online noise \\
    Val & 3{,}829 & 11{,}487 & 7.6\,h & used for all reported val numbers \\
    Test-Clean & 7{,}500 & 22{,}500 & 15.4\,h & seen-style timbres;
    synthesis complete \\
    Test-Acoustic & 7{,}500 & 22{,}493 & --- & 2 held-out timbres +
    unseen noises; synthesis complete (7 missing) \\
    \bottomrule
  \end{tabular}
\end{table}

Every accuracy benchmark in Section~\ref{sec:mainresults} is evaluated on
the \textbf{full validation set (3,829 queries)} --- and confirmed on the
full 7,500-query test split --- with \textbf{true pairwise IoU}; the
2,000-query monitoring subset used during training overestimates Acc@0.5 by
${\sim}5.7$ points and is never used for reported numbers (a
measurement-hygiene note we strongly suggest for this task: several
easy-to-make evaluation errors --- GIoU-derived pseudo-IoU, silent
validation audio, easy subsets --- each inflated our intermediate numbers
before being caught).

\subsection{The Compositional Blind Spot of the Standard Split}
\label{sec:blindspot}

Parsing all (attribute, category) pairs of the standard test split against
the training split under a 30-word attribute lexicon shows that
\textbf{277 of the 278 combinations in the 7,500 test queries also appear in
training} --- the single exception is (``brown'', ``ground track field'').
The strictly compositional (zero-shot combination) subset is therefore
effectively empty: a model can score well on the standard test while having
memorized nearly every attribute--category pair it will ever be asked
about. Acoustic augmentation cannot fix this; it is orthogonal.

\subsection{Audio-DIOR-Comp}
\label{sec:comp}

Following C-GQA~\cite{cgqa2022}, we re-split the corpus at the
\emph{combination} level: parse attribute--category combinations over the
full dataset using a 30-word attribute lexicon; greedily select a held-out
combination set $\mathcal{H}$ under (i) 15--500 queries per combination, (ii)
a total budget of 1,300 queries, (iii) ${\le}3$ held-out combinations per
category and per attribute, and (iv) every retained atom keeps ${\ge}2$
partners on both sides after removal. Atoms of every held-out combination
are amply seen in training --- only their \emph{composition} is unseen
(``blue ship'', ``red car'' are seen; ``blue car'' is not).
Table~\ref{tab:comp} lists the resulting splits.

\begin{table}[t]
  \centering\small
  \caption{Audio-DIOR-Comp re-split. Held-out combinations have atoms seen
  in training; only their composition is unseen.}
  \label{tab:comp}
  \begin{tabular}{@{}lccc@{}}
    \toprule
     & comp-train & comp-val & comp-test \\
    \midrule
    Queries & 26{,}090 & 3{,}698 & 1{,}284 \\
    (attribute, category) combinations & 269 seen & --- &
    \textbf{17 held-out} \\
    \bottomrule
  \end{tabular}
\end{table}

Evaluating on Comp requires re-training on comp-train (the standard model
has seen the held-out compositions). Our first two-arm chain hit a
data-protocol bug --- both arms silently trained on the full train split
instead of comp-train --- and every number from it was discarded and
replaced by the corrected chain below: a supervision-free
baseline and the full-strength recipe (v8: ASR pretraining $+$ transformer
connector $+$ CTC 0.3), both re-trained on comp-train with model selection
on comp-val and final evaluation on comp-test. The preregistered criterion:
if the recipe outperforms the baseline on comp-test by ${\ge}3$ Acc@0.5
(with a difference $<2$ treated as no transfer), the capability gains
measured in Section~\ref{sec:profile} (relation words $+35.7$ from the
collapsed baseline, crossing the frozen front-end's 64.2 inheritance line
at v7's 74.8) and
Section~\ref{sec:retrieval} (sentence discriminability ${\times}15$) have
\emph{transferred to compositional localization}; otherwise we will report
that injection improves the representation --- as the probes certify ---
without transferring to compositional localization.

The measured outcome falls in the no-transfer branch, reported as a parity
result: on the 1,284 comp-test queries the full-strength recipe and the
supervision-free baseline are statistically indistinguishable (Acc@0.5
41.59 vs.\ 40.11, a $+1.48$-point difference within the binomial 95\% CI of
about ${\pm}2.7$ points at $n{=}1{,}284$; mIoU 35.54 vs.\ 34.19;
Table~\ref{tab:compresults} in Section~\ref{sec:compacoustic}), and both
arms lose nearly the same amount from comp-val to comp-test ($-11.69$ and
$-12.21$ Acc@0.5). Under this 20-epoch schedule --- in which the recipe arm
scores 49.10 on the standard validation split, below its converged 64.32
--- no compositional penalty was detected, recombination difficulty is a
property of the task rather than of semantic injection, and compositional
transfer remains open --- now with a controlled measurement.

\subsection{The Acoustic \texorpdfstring{$\times$}{x} Semantic
\texorpdfstring{$2\times2$}{2x2} Protocol}
\label{sec:protocol2x2}

Table~\ref{tab:protocol} defines the four evaluation cells.

\begin{table}[t]
  \centering\small
  \caption{The acoustic $\times$ semantic 2$\times$2 evaluation protocol.}
  \label{tab:protocol}
  \begin{tabular}{@{}lll@{}}
    \toprule
     & Combinations seen & Combinations unseen (Comp) \\
    \midrule
    \textbf{Acoustics seen} (Test-Clean) & standard main table &
    \textbf{Comp@Clean} \\
    \textbf{Acoustics unseen} (Test-Acoustic) & robustness table &
    \textbf{Comp@Acoustic} (hardest cell) \\
    \bottomrule
  \end{tabular}
\end{table}

The four cells separate acoustic robustness (bottom row), semantic
generalization (right column), and their conjunction (bottom-right).
\textbf{Status:} standard main table --- complete
(Section~\ref{sec:mainresults}, with its test-split confirmation);
Comp@Clean --- measured (corrected two-arm chain, Section~\ref{sec:comp}:
the full-strength recipe and the supervision-free baseline are
statistically indistinguishable on the 17 held-out combinations,
Section~\ref{sec:compacoustic}); acoustic-robustness row --- measured
(Section~\ref{sec:compacoustic}); Comp@Acoustic (hardest cell) --- pending.
Currently available cells: the standard main table and its test-split
confirmation (Section~\ref{sec:mainresults}), the Comp@Clean cell, and the
acoustic-robustness row (Section~\ref{sec:compacoustic}).

%% file: experiments.tex
\section{Experiments}
\label{sec:experiments}

\subsection{Setup}
\label{sec:setup}

\begin{itemize}
  \item \textbf{Host.} VLTVG~\cite{vltvg2022} hosts every speech model; the
  speech-side decoder is initialized from the officially released VLTVG
  checkpoint (\texttt{detr-r50-referit.pth}; logged in every run). The text
  upper bound uses the same frozen-encoder/decoder design trained from
  initialization on DIOR-RSVG train (two-stage: 30 epochs $+$ 15-epoch
  fine-tune at $10^{-5}$) --- a disclosed asymmetry: its training schedule
  is longer than the speech models' 30-epoch budget, favoring the
  \emph{upper-bound} row. Published anchors (e.g., VLTVG's 75.79 Acc@0.5)
  are on
  the DIOR-RSVG \emph{test} split under the original protocol (higher
  input resolution, different evaluation script) and are not
  directly comparable to our full-\emph{val} speech-side numbers; the
  remaining ${\approx}4$-point gap between our text route and that anchor
  reflects these protocol differences plus our frozen-BERT setting.
  \item \textbf{Speech models.} Whisper-Base.en (Whisper-Small.en where
  stated) and BERT frozen; adapter, projections, decoder, and heads
  trainable. Batch 4, 640$\times$640 inputs with RandomResize, bf16 AMP,
  AdamW (lr $10^{-4}$ for new modules; $10^{-5}$ for backbone/BERT groups),
  weight decay $10^{-4}$, gradient clip 0.1, single RTX 5060 Laptop 8\,GB.
  Schedules by lineage: the baseline and single-channel runs (v4--v6) train
  30 epochs with step LR ${\times}0.1$ at epochs 10 and 20 and the decoder
  frozen for the first 10 epochs. The v8 lineage trains longer under a
  \textbf{two-stage collapse-stop schedule} that we disclose explicitly: the
  first v8 attempt (constant $10^{-4}$) collapsed at epoch 14 (val Acc@0.5
  2.05\%); it was restarted from the last healthy checkpoint (epoch 11) and
  fine-tuned at $10^{-5}$ (v8\_ft), then extended a second time (v8\_ft2)
  --- ${\approx}47$ epochs total on the same data and supervision, with no
  additional signal beyond the recipe of Section~\ref{sec:channelC}. All
  checkpoint selection uses the validation subset; reported numbers come
  from the full val/test evaluations below.
  \item \textbf{Test audio.} Test-Clean (7,500 queries ${\times}$ 3 voices)
  is fully synthesized and used for the test-split confirmation in
  Section~\ref{sec:mainresults}; Test-Acoustic (7,500 queries, 2 held-out
  timbres, unseen noise clips baked at synthesis) is now synthesized
  (22,493/22,500 clips; 7 missing, an impact of ${\le}0.1$ point) and used
  for the robustness evaluation in Section~\ref{sec:compacoustic}.
  \item \textbf{Metrics.} Acc@0.5 and mIoU~\cite{rstvg2023zhan} with true
  pairwise IoU on the full val set (3,829). Cascade and probe studies use
  the first 500 val queries, identical samples across all compared systems
  (subset Acc@0.5 is easier than full-val and must not be compared across
  tables). For context in Section~\ref{sec:compacoustic}: the same-host
  text route, itself a contaminated upper bound on Comp (trained on the
  standard split, it has seen all 17 held-out combinations and 901 of the
  1,284 comp-test queries), still drops from 73.09 Acc@0.5 on comp-val to
  66.51 on comp-test --- recombination is hard even for text.
  \item \textbf{Relation to~\cite{rstalker2026liu}.} RSTalker's published
  numbers use a different backbone and the unpublished original audio
  corpus; we therefore compare within our controlled family and
  quote~\cite{rstalker2026liu}'s reported numbers only as context,
  marked~$\dagger$.
\end{itemize}

\subsection{Main Results on the Standard Validation Split}
\label{sec:mainresults}

Table~\ref{tab:main} reports the main comparison.

\begin{table}[t]
  \centering\footnotesize\setlength{\tabcolsep}{3pt}
  \caption{Main results on the full Audio-DIOR-R validation set (3,829
  queries, true pairwise IoU). $\Delta$Acc is computed against the speech
  baseline. $^{\dagger}$Reported in~\cite{rstvg2023zhan,rstalker2026liu} on
  their own test protocol, host, and (unpublished) audio corpus --- not
  comparable to our rows; shown only to locate the family.}
  \label{tab:main}
  \begin{tabular}{@{}>{\raggedright\arraybackslash}p{5.35cm}>{\raggedright\arraybackslash}p{2.9cm}lrrr@{}}
    \toprule
    Model & Language supervision & Input & Acc@0.5 & mIoU &
    $\Delta$Acc \\
    \midrule
    Fair text upper bound (same host) & full (paired text) & Text &
    \textbf{72.97} & \textbf{62.31} & $+19.69$ \\
    MGVLF~\cite{rstvg2023zhan}~$^{\dagger}$ & full & Text & 76.78~$^{\dagger}$
    & 78.42~$^{\dagger}$ & --- \\
    RSTalker (reported)~\cite{rstalker2026liu}~$^{\dagger}$ & none & Speech
    & 79.85~$^{\dagger}$ & 78.29~$^{\dagger}$ & --- \\
    Speech baseline & none & Speech & 53.28 & 45.20 & --- \\
    $+$ Channel A: feature mimicry (v4) & mimicry distillation & Speech &
    51.68 & 44.29 & $\mathbf{-1.60}$ \\
    $+$ Channel B: CTC word-level from scratch (v5) & word-level & Speech &
    53.72 & 45.64 & $+0.44$ \\
    $+$ Channel C: ASR pretraining $+$ CTC 0.3 (v6) & word-level, two-stage
    & Speech & 53.85 & 45.83 & $+0.57$ \\
    $+$ Full-strength recipe, Base.en, 30-epoch budget (v7) & all channels,
    matched budget & Speech & 52.68 & 45.20 & $-0.60$ \\
    $+$ Full recipe: Whisper-Small $+$ extended two-stage training (v8\_ft2)
    & two-stage scaled & Speech & \textbf{64.32} &
    \textbf{54.85} & $\mathbf{+11.04}$ \\
    \bottomrule
  \end{tabular}
\end{table}

\textbf{Test-split confirmation.} The same seven systems evaluated on the
full 7,500-query standard test split (Test-Clean audio,
Section~\ref{sec:audiodiorr}) reproduce the val ordering and the recipe's
gain (Table~\ref{tab:testconf}).

\begin{table}[t]
  \centering\small\setlength{\tabcolsep}{4pt}
  \caption{Test-split confirmation: the full 7,500-query standard test
  split (Test-Clean audio, Section~\ref{sec:audiodiorr}), same systems as
  Table~\ref{tab:main}.}
  \label{tab:testconf}
  \begin{tabular}{@{}lrr@{}}
    \toprule
    Model & Acc@0.5 & mIoU \\
    \midrule
    Fair text upper bound (same host) & \textbf{71.73} & \textbf{61.59} \\
    Speech baseline & 51.75 & 44.39 \\
    $+$ Channel A: mimicry (v4) & 51.28 & 43.82 \\
    $+$ Channel B: CTC (v5) & 52.19 & 44.98 \\
    $+$ Channel C: ASR pretraining (v6) & 52.53 & 45.06 \\
    $+$ Full strength intermediate (v7) & 52.55 & 44.70 \\
    $+$ Full recipe final (v8\_ft2) & \textbf{62.81} & \textbf{53.96} \\
    \bottomrule
  \end{tabular}
\end{table}

\begin{figure}[t]
  \centering
  \includegraphics[width=0.62\textwidth]{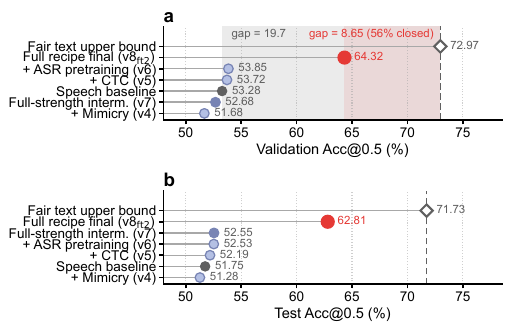}
  \caption{The recipe gain and its test-split reproduction.
  \textbf{a}, Full-validation Acc@0.5 for the seven systems of
  Table~\ref{tab:main}: the two-stage recipe (v8\_ft2, red) closes 56\% of
  the 19.7-point gap to the fair text upper bound (8.65 remains).
  \textbf{b}, The same systems on the 7,500-query test split
  (Table~\ref{tab:testconf}). The single-channel variants (v4--v6) sit
  within evaluation noise of the baseline at matched budgets (v4--v7).}
  \label{fig:mainresults}
\end{figure}

The $+11.06$ test-side gain over the baseline (62.81 vs.\ 51.75) matches
the val-side $+11.04$, ruling out a single-split artifact behind the
recipe's improvement (Fig.~\ref{fig:mainresults}).

\paragraph{Reading.} (i) The fair text upper bound is 72.97/62.31. (ii) On
standard splits and at the matched 30-epoch budget, \emph{every} injection
channel is within evaluation noise of the baseline ($-1.60$ to $+0.57$,
v4--v7): the box loss itself already organizes category semantics (baseline
probe 31.0 vs.\ chance 5.0), so the channels add no behavioral gain at this
granularity. (iii) \textbf{Scaling the speech front-end to Whisper-Small
and extending training under the full two-stage recipe move the speech
route to 64.32 ($+11.04$ over the matched baseline), closing 56\% of the
19.7-point text--speech gap.} We attribute this gain to the \emph{joint}
effect of front-end capacity, extended schedule, and ASR pretraining ---
the matched-budget cells above bound the channels' standalone contribution
to within noise, and the missing capacity/schedule controls (Whisper-Small
baseline, longer-budget channels) are listed as future work
(Section~\ref{sec:conclusion}). The remaining 8.65 points appear structural
(word-identity recovery capacity); see the capability profile
(Section~\ref{sec:profile}) and the plateau analysis below.

\paragraph{Plateau note (v8\_ft2).} Extending v8 fine-tuning for
${\approx}20$ further epochs (19 logged) under a constant low learning rate
yields a high plateau (validation subset 67.9 $\to$ 69.25 at epoch 14, then
oscillation at 66.8--69.3 while training loss keeps decreasing): the model
is still learning, but no longer learning \emph{localization} --- the
semantics recoverable from the audio front-end appear exhausted at this
capacity. This plateau is our strongest evidence that the remaining gap is
structural (front-end capacity), not an artifact of under-training or
evaluation noise; confirming it requires the \{Base, Small\}${\times}$%
longer-training control we list as future work
(Section~\ref{sec:conclusion}).

\subsection{Direct Alignment vs.\ a Domain-Trained ASR Cascade}
\label{sec:directcascade}

The anti-ASR premise of~\cite{rstalker2026liu} --- transcription errors
propagate catastrophically --- has never been quantified with a controlled
cascade family. We build two in-domain cascades from the channel-C ASR
pretraining of Section~\ref{sec:channels} (greedy-decoded transcripts fed
to the \emph{same host's text path}): the v1-generation recognizer (conv
adapter, CTC 0.53) and the v3-generation recognizer (transformer
connector, decoder-init, CTC 0.029). Four routes, identical 500 val
samples (Table~\ref{tab:cascade}).

\begin{table}[t]
  \centering\small
  \caption{Direct alignment vs.\ domain-trained ASR cascades vs.\ the text
  upper bound (first 500 val queries, identical samples; all routes in the
  same harness). The direct--baseline row is the 30-epoch matched-budget
  baseline; the direct--full-recipe row is v8\_ft2 (the strongest direct
  model), evaluated in the same harness. The v1 rows are from the original
  run; its direct and text routes reproduced at 61.20/52.34 and
  81.00/68.97 in the v3 run.}
  \label{tab:cascade}
  \begin{tabular}{@{}lrrr@{}}
    \toprule
    Route & Input & Acc@0.5 & mIoU \\
    \midrule
    ASR cascade, v1 recognizer (CTC 0.53) & transcript $\to$ text path & 53.00 &
    46.34 \\
    \textbf{ASR cascade, v3 recognizer (CTC 0.029)} & transcript $\to$ text path &
    \textbf{80.80} & \textbf{68.93} \\
    Direct speech (matched-budget baseline) & speech $\to$ shared decoder &
    61.20 & 52.34 \\
    Direct speech (full recipe, v8\_ft2) & speech $\to$ shared decoder &
    72.00 & 61.27 \\
    Text upper bound (ground-truth expressions) & text & 81.00 & 68.97 \\
    \bottomrule
  \end{tabular}
\end{table}

\begin{figure}[!htb]
  \centering
  \includegraphics[width=0.6\textwidth]{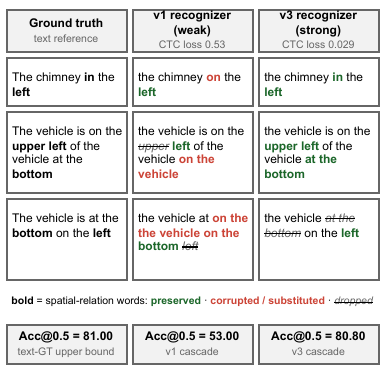}
  \caption{Cascade transcript analysis on identical queries: the
  v1-generation recognizer (CTC 0.53) corrupts or drops the
  spatial-relation words that select the target, while the v3-generation
  recognizer (CTC 0.029) preserves them --- and the v3 cascade matches the
  text upper bound (80.80 vs.\ 81.00 Acc@0.5) where the v1 cascade loses
  28.0 points.}
  \label{fig:cascade}
\end{figure}

\textbf{Finding (i), as preregistered and measured: the direct-vs-cascade
question is decided by recognizer quality, and the strong cascade still
wins --- but by 8.8 points over the strongest direct model, not 19.6.}
Against the weak v1 cascade, the matched-budget direct baseline wins by
$+9.0$ Acc@0.5 in the original harness (62.00 vs.\ 53.00; reproduced at
61.20 vs.\ 53.00, $+8.2$) --- transcription errors concentrate on
\textbf{spatial-relation words} (\emph{upper-left}, \emph{bottom},
\ldots) --- precisely the tokens that determine which of several visually
similar instances is the target; word identity survives the cascade, word
\emph{role} does not. But against the strong v3 cascade --- whose
transcripts are near-perfect, with the same relation words intact --- the
ordering reverses: the cascade reaches \textbf{80.80, statistically
indistinguishable from the text upper bound (81.00)}, and beats direct
alignment by $+8.8$ points over the \emph{strongest} direct model
(72.00) and $+19.6$ over the matched-budget baseline. The anti-ASR premise
therefore holds only
against \emph{weak} recognizers; with a domain-trained recognizer at CTC
0.029 (closed vocabulary, 81k in-domain expressions --- exactly what
channel C's pretraining provides), the cascade recovers the text path's
performance in full. The residual case for direct alignment is elsewhere:
it avoids the recognizer entirely (deployment simplicity, no
transcription exposure) and remains the only route when in-domain ASR
training data does not exist. This reversal is reported as measured, per
the criterion preregistered before the v3 run (a strong cascade
$\ge 62.0$ rewrites the finding). Fig.~\ref{fig:cascade} inspects the
transcripts behind both comparisons.

\subsection{Semantic Capability Profile}
\label{sec:profile}

Table~\ref{tab:probes} reports linear probes on the pooled speech
representation (500 matched val queries; columns span the supervision dose:
baseline = no language supervision; v4 = mimicry; v5 = CTC-only; v6 = ASR
pretraining $+$ CTC 0.3 (conv adapter); v7 = the full-strength recipe at
the Base.en stage (30-epoch budget, \texttt{connector\_v2} pretraining);
v8 = full recipe final (small.en, layer-sum, decoder-init,
\texttt{connector\_v3}, extended two-stage training)).
Table~\ref{tab:buckets} adds bucketed localization accuracy.

\begin{table}[t]
  \centering\footnotesize\setlength{\tabcolsep}{3.5pt}
  \caption{Linear semantic probes on the pooled speech representation (500
  matched val queries); columns span the supervision dose.}
  \label{tab:probes}
  \begin{tabular}{@{}lrrrrrrr@{}}
    \toprule
    Probe dimension & Chance & Baseline & v4 (mimicry) & v5 (CTC) &
    v6 (ASR-pre) & v7 (full) & v8 (final) \\
    \midrule
    Category word (20-way) & 5.0 & 31.0 & 33.0 & \textbf{53.0} & 34.5 &
    \textbf{66.0} & 41.5 \\
    Attribute word (9-way, attr subset) & 12.5 & 34.1 & \textbf{41.5} &
    36.6 & 39.0 & 31.7 & 31.7 \\
    Relation word (8-way, rel subset) & 20.0 & 39.1 & 53.6 & 57.0 & 62.3 &
    \textbf{74.8} & 60.9 \\
    Bag-of-words (top-40, multi-label) & ${\approx}15.8$ & 88.1 & 86.5 &
    88.1 & 88.1 & 88.1 & 87.3 \\
    \bottomrule
  \end{tabular}
\end{table}

\begin{table}[t]
  \centering\footnotesize\setlength{\tabcolsep}{2.5pt}
  \caption{Bucketed localization (Acc@0.5); bucket sizes $n = 376$ / 60 /
  64 for relation / attribute / simple --- the attribute and simple buckets
  are small, so we read them as trends and lean on the relation bucket for
  conclusions.}
  \label{tab:buckets}
  \begin{tabular}{@{}lrrrrrr@{}}
    \toprule
    Query bucket & Baseline & v4 & v5 (CTC) & v6 & v7 (full) &
    \textbf{v8 (final)} \\
    \midrule
    Relation-bearing & 60.9 & 59.0 & 63.0 & 60.9 & 59.8 & \textbf{70.5} \\
    Attribute-bearing & 56.7 & \textbf{58.3} & 56.7 & 55.0 & 55.0 &
    \textbf{68.3} \\
    Simple (category-only) & 71.9 & 65.6 & 65.6 & 70.3 & 68.8 &
    \textbf{76.6} \\
    \bottomrule
  \end{tabular}
\end{table}

\paragraph{Probe controls (preregistered).} To decide whether word-class
decodability is grounding-taught or inherited from the frozen front-end,
we probe three untrained representations under the identical protocol
(same 500 samples, 60/40 train/test split; probe test sets ${\approx}150$
relation / ${\approx}24$ attribute / ${\approx}26$ simple queries ---
single probe fit per cell, no seed replication; the attribute and simple
cells are therefore directional only):

\begin{table}[h]
  \centering\small
  \caption{Untrained probe controls: word-class decodability is already
  present in the frozen front-end.}
  \label{tab:probecontrols}
  \begin{tabular}{@{}lrrrr@{}}
    \toprule
    Probe (chance) & Frozen Whisper & Random adapter & Random proj. & Trained baseline \\
    \midrule
    Category 20-way (5.0) & 55.5 & 50.0 & 52.0 & 31.0 \\
    Attribute 9-way (12.5) & 41.5 & 41.5 & 34.1 & 34.1 \\
    Relation 8-way (20.0) & \textbf{64.2} & 58.9 & 58.3 & 39.1 \\
    \bottomrule
  \end{tabular}
\end{table}

\begin{figure}[t]
  \centering
  \includegraphics[width=0.9\textwidth]{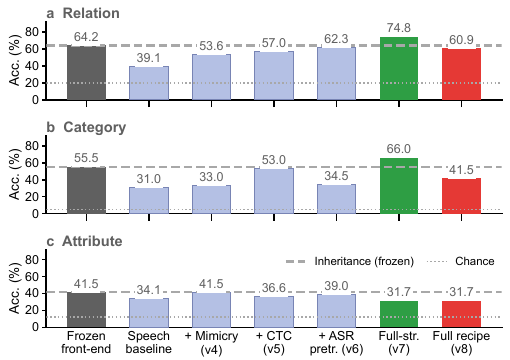}
  \caption{Word-class decodability is inherited from the frozen front-end
  and recovered by the distillation channels.
  \textbf{a}, Relation-word probe (the hero): conventional grounding
  collapses the inheritance (39.1 vs.\ 64.2); the channels recover it
  monotonically and only v7 (74.8, green) crosses the inheritance line
  (gray dashed). \textbf{b}, Category probe. \textbf{c}, Attribute probe.
  Dotted line = chance; dark neutral bar = frozen front-end (the reference).
  v8 (red) trades probe decodability for localization
  (Fig.~\ref{fig:tsne}, Table~\ref{tab:buckets}).}
  \label{fig:recovery}
\end{figure}

\textbf{The verdict is unambiguous: word-class decodability is an
inheritance of the frozen front-end, not a product of grounding.} With no
grounding training whatsoever, the frozen Whisper features decode
category/attribute/relation words at 55.5/41.5/64.2 --- at or above every
trained model in the family except v7's relation score. Conventional
grounding training \emph{collapses} this inheritance (baseline relation
39.1 vs.\ frozen 64.2, $-25.1$); the word-class ``dose response'' below is
therefore a \emph{recovery curve} against the inheritance line, not a
creation curve against chance.

\begin{figure}[!htb]
  \centering
  \includegraphics[width=0.72\textwidth]{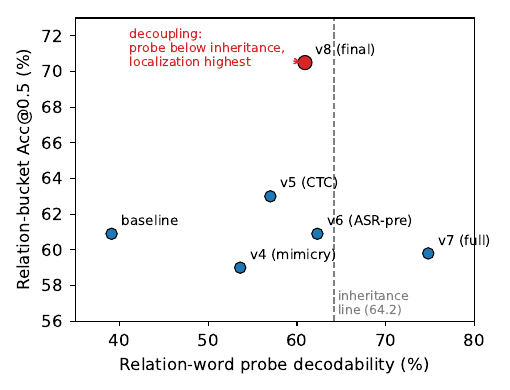}
  \caption{Probe decodability vs.\ localization across the six systems
  (relation-word probe vs.\ relation-bucket Acc@0.5; Table~\ref{tab:probes}
  and Table~\ref{tab:buckets}). The full-strength v8 sits below the frozen
  front-end's inheritance line (dashed) on the probe yet leads every bucket
  --- injected semantics sink from probe-detectable representation into
  localization behavior.}
  \label{fig:probeloc}
\end{figure}

\paragraph{Findings.}
\begin{enumerate}
  \item \textbf{The front-end already knows the words --- grounding training
  collapses the evidence.} With no grounding training at all, the frozen
  front-end decodes category/attribute/relation words at
  55.5/41.5/64.2 (Table~\ref{tab:probecontrols}): the unsupervised
  baseline's 31.0/34.1/39.1 is not semantics created by box regression but
  the front-end's inheritance surviving grounding training in reduced
  linear form. Word-class probes measure how much inherited semantics a
  recipe \emph{preserves}, not what it creates.
  \item \textbf{The dose response is a recovery curve --- and the full
  recipe crosses the inheritance line} (Fig.~\ref{fig:recovery}). Read against the frozen front-end's
  64.2 rather than chance 20.0, the relation-word sequence 39.1 $\to$ 53.6
  (mimicry) $\to$ 57.0 (CTC) $\to$ 62.3 (ASR pretraining, conv adapter)
  $\to$ \textbf{74.8 (full strength)} means: conventional grounding gives up
  25.1 points of inherited decodability; the word-level recipes buy it back
  ($+23.2$ through the v6 dose axis, break-even at v6), and the
  full-strength v7 \emph{exceeds} the front-end by $+10.6$ --- the only
  state in the family where training adds relation-word decodability beyond
  inheritance. Sentence-level discriminability, by contrast, is genuinely
  from-nothing (Section~\ref{sec:retrieval}).
  \item \textbf{Decodability gains do not automatically become grounding
  gains --- but the final recipe closes the loop.} v4 raises
  attribute/relation decodability while \emph{losing} 1.6 Acc@0.5; v6
  raises both. The final model (v8, full strength $+$ extended training)
  shows the \emph{decoupling} explicitly: its probe scores (category 41.5,
  relation 60.9) sit below both v7's (66.0/74.8) and the frozen front-end's
  (55.5/64.2), yet its bucketed localization
  leads every column --- relation 70.5 ($+9.6$ over the baseline,
  $n{=}376$ --- the solid evidence), with small-sample positive trends on
  attribute 68.3 ($+11.6$, $n{=}60$) and simple 76.6 ($+4.7$, $n{=}64$).
  \textbf{When word-identity recovery is strong enough, injected semantics
  sink from ``probe-detectable representation'' into ``localization
  behavior''} --- the state a deployed system actually needs
  (Fig.~\ref{fig:probeloc}).
  \item \textbf{Inherited decodability is the rule --- probes must always be
  controlled.} Bag-of-words is ${\sim}88\%$ decodable
  in the unsupervised baseline and \emph{does not move across any of the
  six models} (88.1 / 86.5 / 88.1 / 88.1 / 88.1 / 87.3); the word-class
  controls above extend the same Johnson--Lindenstrauss caution from word
  identity to the word-class probes themselves. \textbf{Probe dose
  responses in this paper are therefore reported against two nulls ---
  chance and the frozen front-end}; only the v7 relation score (74.8)
  clears both.
\end{enumerate}

\begin{figure}[!htb]
  \centering
  \includegraphics[width=0.65\textwidth]{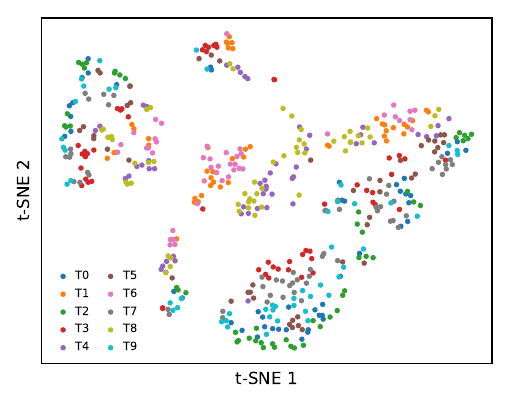}
  \caption{t-SNE of the v4 orthogonal-residual features (500 val queries,
  colored by the 10 synthesis timbres): timbre structure survives in the
  reserved subspace (fresh-probe decodability 69.0\%, chance 10\%), while
  collapsing to near chance in grounding-trained pooled representations
  without the reservation (Section~\ref{sec:retrieval}).}
  \label{fig:tsne}
\end{figure}

\subsection{Sentence-Level Discriminability}
\label{sec:retrieval}

An earlier-protocol probe run (intermediate checkpoint of the v4 run;
20-way category probe with timbre/noise shortcut excluded) measured
\textbf{82.5\% category decodability (chance 5\%)} and
speech$\,\to\,$text retrieval (Table~\ref{tab:retrieval}).

\begin{table}[t]
  \centering\small
  \caption{Speech$\,\to\,$paired-text retrieval (earlier-protocol probe
  run, v4 checkpoint at epoch 21).}
  \label{tab:retrieval}
  \begin{tabular}{@{}lrr@{}}
    \toprule
    Retrieval (speech $\to$ paired text) & R@1 & R@10 \\
    \midrule
    Chance & ${\approx}0.2$ & ${\approx}0.2$ \\
    Speech baseline & 0.4 & 1.8 \\
    Mimicry-distilled (v4, ckpt ep21) & \textbf{2.6} & \textbf{26.6} \\
    \bottomrule
  \end{tabular}
\end{table}

Without language supervision, the pooled speech embedding carries almost no
sentence-level identity (R@1 0.4\% ${\approx}$ chance); even the weak
mimicry channel lifts R@10 roughly 15-fold (1.8 $\to$ 26.6). Sentence
discriminability is a \emph{from-nothing} capability that only language
supervision provides.

\paragraph{Speaker / timbre preservation (preregistered probe, chance
10\%).} The orthogonal subspace of Section~\ref{sec:channelA} anchors
acoustic identity through grounding training. The cross-model comparable
number is the pooled representation: the mimicry model (v4) retains
\textbf{51.0\%} timbre decodability, while purely grounding-trained
representations --- where nothing supervises acoustic identity --- collapse
to near chance (baseline 14.0\%, full-recipe v8 13.5\%). The residual
subspace explains the difference: the \emph{trained} speaker probe reads
synthesis timbre from the orthogonal residual alone at 75.0\% (fresh
post-hoc probe: 69.0\%). We note the boundary of this claim: timbre
preservation did not translate into robustness here --- on the two
held-out Test-Acoustic timbres, v4 scores 50.48 vs.\ the baseline's 51.00
(par; Section~\ref{sec:compacoustic}) --- so the reservation is currently
an observation about representational structure, not a demonstrated
robustness mechanism, and it is timbre-granularity evidence; emphasis-level
prosody remains unmeasured (Section~\ref{sec:conclusion}).

\subsection{Why Feature Mimicry Fails: Loss-Share Accounting}
\label{sec:gradient}

The $-1.60$ of channel A is not bad luck; it is arithmetic. Logging the
loss terms during v4 training (we report loss-magnitude shares; a
gradient-norm accounting is future work) yields Table~\ref{tab:lossshare}.

\begin{table}[t]
  \centering\small
  \caption{Loss-share accounting during v4 training (loss-magnitude
  shares).}
  \label{tab:lossshare}
  \begin{tabular}{@{}llp{4.2cm}@{}}
    \toprule
    Loss family & Typical magnitude & Share of training loss \\
    \midrule
    Box losses ($\ell_1$ $+$ GIoU, deep supervision, teacher stream) &
    20--40 & ${\sim}98$--$99\%$ \\
    Distillation terms
    $L_q + L_{\mathrm{attn}} + L_{\mathrm{sem}} + L_{\mathrm{orth}}$ &
    ${\approx}0.8$ (measured) & \textbf{1--2\%} \\
    CTC term (channel B, $\lambda=1.0$) & --- &
    \textbf{${\approx}35\%$ at initialization, 3--8\% at convergence} \\
    \bottomrule
  \end{tabular}
\end{table}

At conventional weights, mimicry is a \textbf{numerical no-op}:
${\pm}1$-point Acc@0.5 differences against the baseline are within
evaluation noise (single-seed runs; the binomial 95\% CI at $n=3{,}829$ is
about ${\pm}1.6$ points). The representation still shifts (relation
decodability 39.1$\,\to\,$53.6), but not enough, and not in a way that
reaches the localization pathway. The design lesson generalizes beyond this
paper: \emph{an auxiliary semantic loss must be ranked by the share of the
training loss it actually commands, not by the elegance of its target} ---
a criterion aligned with multi-task learning practice, where interventions
are judged by their actual gradient
contributions~\cite{pcgrad2020yu,gradnorm2018chen} --- which immediately
explains why word-level CTC
(Eq.~\eqref{eq:lctc}), whose gradients flow through every token, is the
channel that works: its share exceeds channel A's by an order of magnitude
at initialization and remains several times larger after convergence. The
mimicked feature itself was verified healthy in the earlier-protocol probe
run of Section~\ref{sec:retrieval} (category probe 82.5\%, inter-token
cosine 0.347 vs.\ BERT's 0.589), ruling out representation homogenization
as the failure cause.

\paragraph{What channel A buys, and what it cannot buy at this weight.}
The controls of Section~\ref{sec:profile} sharpen the verdict on feature
mimicry from ``failed'' to \emph{two real representation-level effects with
one behavioral shortfall}. (i)~\textbf{Loss-efficient restoration}: mimicry
alone recovers $+14.5$ relation-word points (39.1\,$\to$\,53.6) while
commanding only 1--2\% of the training loss --- per unit of commanded
signal, the most efficient semantic restorer in the family.
(ii)~\textbf{Modality preservation}: its orthogonal reservation keeps
acoustic identity decodable through grounding training (timbre 75.0/69.0
from the residual, vs.\ 14.0/13.5 ${\approx}$ chance for grounding-only
representations; Section~\ref{sec:retrieval}). (iii)~The shortfall is
behavioral: 1--2\% of the loss is not enough magnitude to move Acc@0.5.
The design conclusion is therefore \emph{not} ``drop distillation'' --- it
is that geometry-level mimicry and word-level magnitude buy different
currencies, and a deployable recipe needs the latter
(Section~\ref{sec:mainresults}) while the former is what keeps the
student's semantics anchored and its speaker identity intact.

\subsection{Compositional and Acoustic Generalization}
\label{sec:compacoustic}

The protocol is defined in Section~\ref{sec:protocol2x2}. Three of the
four cells are now measured.

\paragraph{Comp@Clean (corrected two-arm chain,
Section~\ref{sec:comp}).} Both arms were re-trained on comp-train (26,090
queries) under identical 20-epoch schedules, with model selection on the
comp-val monitoring subset (best epoch 19 in both arms; monitoring-subset
numbers are optimistic and must not be compared across tables) and final
evaluation on the 1,284-query comp-test with a merged three-provenance
audio bundle (Table~\ref{tab:compresults}). The two arms are statistically
indistinguishable: $+1.48$ Acc@0.5 / $+1.35$ mIoU, within the binomial
95\% CI of about ${\pm}2.7$ points at $n{=}1{,}284$ --- the preregistered
$<2$-point branch applies. Both arms lose nearly the same amount from
comp-val to comp-test ($-11.69$ and $-12.21$ Acc@0.5), indicating that
recombination difficulty is a property of the task, not of semantic
injection. One scope note: at this 20-epoch schedule the recipe arm scores
49.10 on the standard validation split --- below its converged 64.32 --- so
the parity claim is confined to that schedule: \emph{no compositional
penalty was detected under a 20-epoch schedule}, gains localize to seen
compositions at convergence, and compositional transfer remains open ---
now with a controlled measurement.

\begin{table}[t]
  \centering\small
  \caption{Audio-DIOR-Comp corrected two-arm chain (Comp@Clean cell). Both
  arms re-trained on comp-train (26,090 queries) under identical 20-epoch
  schedules, model selection on the comp-val monitoring subset (best epoch
  19 in both arms), evaluation on the 1,284-query comp-test with a merged
  three-provenance audio bundle. The $+1.48$ Acc@0.5 difference is within
  the binomial 95\% CI (${\approx}{\pm}2.7$ points at $n{=}1{,}284$): the
  preregistered $<2$-point branch applies --- the two arms are
  statistically indistinguishable.}
  \label{tab:compresults}
  \begin{tabular}{@{}>{\raggedright\arraybackslash}p{5.9cm}rrr@{}}
    \toprule
    Arm & comp-val subset best & comp-test Acc@0.5 & comp-test mIoU \\
    \midrule
    Supervision-free baseline (conv adapter, Whisper-Base.en) & 51.80 &
    40.11 & 34.19 \\
    Full-strength recipe (transformer connector, Whisper-Small.en) & 53.80 &
    41.59 & 35.54 \\
    \bottomrule
  \end{tabular}
\end{table}

\paragraph{Acoustic robustness (Test-Acoustic).} The standard-split speech
baseline and the full-recipe model (v8\_ft2) evaluated on the 7,500-query
Test-Acoustic split (2 held-out timbres, unseen noise clips baked at
synthesis): the baseline scores 51.00/44.01 against its Test-Clean
51.75/44.39 ($-0.75$), and v8\_ft2 scores 62.67/53.89 against its
Test-Clean 62.81/53.96 ($-0.14$). The recipe's gain is fully preserved
under unseen acoustics: $+11.67$ Acc@0.5 on Test-Acoustic vs.\ $+11.06$ on
Test-Clean. (Test-Acoustic synthesis produced 22,493/22,500 clips; the 7
missing clips affect evaluated totals by at most 0.1 point.)

\paragraph{Comp@Acoustic (hardest cell):} pending.

(Preregistered criterion as stated in Section~\ref{sec:comp}: ${\ge}3$ pt
$\to$ transfer claim; $<2$ pt $\to$ representation-only claim; all results
reported as measured, including if null.)

\subsection{Inference Cost}
\label{sec:cost}

All semantic machinery (text stream, teacher losses, CTC head, subspace
$\mathbf{P}$, probes) exists only at training time; at inference the model
is a speech-only grounder of the identical host architecture --- matching
the cost of a speech-only grounder with the \emph{same front-end}
(Whisper-Small for v8\_ft2, Whisper-Base for the baseline). Measured
latency/params on the deployment GPU:~\TBD{} (to be measured).

\subsection{Qualitative Results}
\label{sec:qualitative}

Fig.~\ref{fig:teaser} (Introduction) condenses the task, the dual-stream
host, and the headline result. Fig.~\ref{fig:framework}
(Section~\ref{sec:host}) presents the dual-stream
host and the three injection channels. Fig.~\ref{fig:cascade} (Sec.\
V-C) shows the
cascade transcript analysis underlying Finding (i);
Fig.~\ref{fig:mainresults} visualizes the main-table gain and its test-split
reproduction; Fig.~\ref{fig:probeloc}, Fig.~\ref{fig:recovery}, and
Fig.~\ref{fig:tsne} visualize the capability findings from logged data.

%% file: conclusion.tex
\section{Conclusion and Limitations}
\label{sec:conclusion}

We presented SemEcho, an anatomy of semantic supervision for speech-driven
remote sensing visual grounding. Its instrument --- a VLTVG dual-stream
host where speech and text share one decoder --- yields the missing fair
text upper bound (72.97 Acc@0.5 / 62.31 mIoU), and its measurement suite
(linear semantic probes, a matched-sample cascade comparison, and
loss-share accounting) turns ``does the audio grounder understand
language?'' into answerable sub-questions. The answers: the
direct-vs-cascade question is decided by recognizer quality --- direct
alignment beats a weak domain cascade by 9.0 points (transcription errors
concentrate on spatial-relation words), while a strong domain cascade
(CTC 0.029) reaches the text upper bound and beats direct by 19.6; word-class
decodability is inherited from the frozen front-end (55.5/41.5/64.2) and
collapsed by conventional grounding training (31.0/34.1/39.1) --- recipes
are judged by how much inheritance they preserve or restore, and only
sentence-level discriminability is learned from nothing; and semantic injection works
only when its share of the training loss is commensurate with the task
loss --- feature mimicry at conventional weights does not move Acc@0.5
(1--2\% of the training loss, $-1.60$ Acc@0.5), yet is the most
loss-efficient semantic restorer ($+14.5$ relation points at that share)
and the only channel that anchors speaker identity through grounding
training (51.0\% pooled timbre decodability vs.\ ${\approx}$chance without;
75.0\% from its
orthogonal residual), while word-level CTC
supervision and two-stage ASR pretraining restore relation-word decodability
monotonically by $+23.2$ points (through the v6 dose axis; the
full-strength v7 crosses the inheritance line at 74.8) while leaving the
standard split within noise of the baseline ($+0.57$). Scaling the speech front-end
to Whisper-Small under the full two-stage recipe then moves the speech
route to 64.32 Acc@0.5 / 54.85 mIoU ($+11.04$ over the matched baseline,
reproduced on the 7,500-query test split at 62.81), closing 56\% of the
19.7-point text--speech gap --- a joint effect of front-end capacity,
extended schedule, and ASR pretraining; the gain persists without acoustic
penalty
(62.67 vs.\ 51.00 Acc@0.5 on the held-out-timbre Test-Acoustic split,
$+11.67$ vs.\ $+11.06$ on clean audio) and no compositional penalty was
detected
(on the 17 held-out combinations the full recipe and the supervision-free
baseline are statistically indistinguishable, 41.59 vs.\ 40.11, under a
20-epoch schedule), although
compositional transfer itself remains open; the remaining 8.65 points
appear structural
--- a low-learning-rate plateau persists while training loss keeps
falling, and injected semantics sink from probe-detectable representation
into localization behavior (bucketed Acc@0.5 improves on every query type
--- solidly on the relation bucket, $n{=}376$, with small-sample positive
trends elsewhere --- even where probe scores saturate). Alongside the
anatomy we release Audio-DIOR-R (80,973 clips, open synthesis pipeline) and
the Audio-DIOR-Comp protocol with a 2$\times$2 acoustic$\times$semantic
matrix.

\paragraph{Limitations.}
(i) All speech is synthetic (TTS); real-field recording remains future
work, as in~\cite{rstalker2026liu}. (ii) Horizontal boxes only. (iii) Our
single 8-GB-GPU budget enforces 640-px training with frozen encoders and
modest schedules (30 epochs for the single-channel runs; ${\approx}47$
epochs across the full recipe's disclosed two-stage collapse-stop schedule,
Section~\ref{sec:setup}) --- absolute numbers will move with stronger
compute, though the \emph{relative} anatomy (probe dose response, cascade
deficit, loss shares) is budget-robust. (iv) Probes measure linear
\emph{decodability}, not understanding; we use them comparatively. (v)
Prosody evidence is at timbre granularity; emphasis-level minimal pairs are
planned. (vi) The acoustic-robustness row and the Comp@Clean cell are now
measured (Section~\ref{sec:compacoustic}) --- the recipe's gain is fully
preserved under unseen acoustics, and the recipe and the supervision-free
baseline are statistically indistinguishable on held-out compositions; only
the Comp@Acoustic cell remains pending and will be reported as measured,
including if null.

Future work extends the probe-driven anatomy to multilingual queries,
rotated boxes, and speech-instructed remote sensing agents --- where the
ear must understand before the agent can act.